\documentclass[letterpaper, 10 pt, conference]{ieeeconf}  

\IEEEoverridecommandlockouts                              
\newif\ifanon
\anonfalse  

\title{PM-Nav: Priori-Map Guided Embodied Navigation in Functional Buildings}

\ifanon
  \author{Anonymous Authors%
    \thanks{Paper under double-blind review.}%
  }
\else
  \author{%
    Jiang Gao$^{1,2*}$,
    Xiangyu Dong$^{2*}$, 
    Haozhou Li$^{2}$,
    Haoran Zhao$^{3}$,
    Yaoming Zhou$^{3}$,
    Xiaoguang Ma$^{1,2\dagger}$
    \thanks{$^{1}$ The authors are with the Faculty of Robot Science and Engineering at Northeastern University, Shenyang, China}%
    \thanks{$^{2}$ The authors are with the Foshan Graduate School of Innovation at Northeastern University, Foshan, China}%
    \thanks{$^{3}$ The authors are with the School of Aeronautic Science and Engineering at Beihang University, Beijing, China}%
    \thanks{$^{*}$ These authors contributed equally to this work.}%
    \thanks{$^{\dagger}$ Corresponding author: Xiaoguang Ma (maxg@mail.neu.edu.cn)}%
  }
\fi

\usepackage{graphicx}
\usepackage{times}
\usepackage{helvet}
\usepackage{courier}
\usepackage[hyphens]{url}
\usepackage{cite}
\usepackage{hyperref}
\usepackage{caption}
\usepackage{algorithm}
\usepackage{algorithmic}
\usepackage{amsmath}
\usepackage{tabularx}
\usepackage{booktabs}
\usepackage{multirow}

\begin{document}

\maketitle
\thispagestyle{empty}
\pagestyle{empty}

\begin{abstract}
Existing language-driven embodied navigation paradigms face challenges in functional buildings (FBs) with highly similar features, as they lack the ability to effectively utilize priori spatial knowledge. To tackle this issue, we propose a Priori-Map Guided Embodied Navigation (PM-Nav), wherein environmental maps are transformed into navigation-friendly semantic priori-maps, a hierarchical chain-of-thought prompt template with an annotation priori-map is designed to enable precise path planning, and a multi-model collaborative action output mechanism is built to accomplish positioning decisions and execution control for navigation planning. Comprehensive tests using a home-made FB dataset show that the PM-Nav obtains average improvements of 511\% and 1175\%, and 650\% and 400\% over the SG-Nav and the InstructNav in simulation and real-world, respectively. These tremendous boosts elucidate the great potential of using the PM-Nav as a backbone navigation framework for FBs.
\end{abstract}

\section{INTRODUCTION}
Current research on language-driven embodied navigation mainly focuses on small-scale indoor scenes with enclosed spaces and simple structures. The core of such research lies in parsing natural language instructions, perceiving the surrounding environment, and completing tasks such as exploration, decision-making, and target localization\cite{irshad2021hierarchical}, \cite{tan2022depth}, \cite{yokoyama2024vlfm}, \cite{long2024discuss}, \cite{zhu2025minivln}, \cite{qiao2025open}. Functional buildings (FBs) feature a standardized spatial layout and are designed to meet the functional needs of specific organizations, such as schools, hospitals, and governments. Instead of focusing on personalized or artistic expression, they feature open spaces, complex structures, and highly similar indicative features. In such environments, traditional navigation methods perform poorly. Firstly, existing research in the path planning phase mostly relies on structured discrete nodes or simplified maps, lacking the capability for global planning in complex FBs. Secondly, current vision-language models (VLMs) have significant limitations in image logical reasoning, unable to accurately understand the spatial logical relationships which exist in complex priori maps. This makes it difficult to effectively convert priori spatial information into navigation guidance. Finally, there are numerous interferences and ambiguous perceptual cues in complex FB environments, making it hard for intelligent agents to determine the next action in the stage of action decision-making.


Although navigation tasks in FBs hold significant practical value, such as for service robots, related research is barely available. In comparison, humans can successfully navigate in FBs using priori maps. Inspired by this, we design a priori-map guided navigation to help the model gain deeper understanding of the spatial structure of FBs, significantly enhance navigation reasoning capabilities and execute precise decisions.

Although VLMs are highly expected to excel in map understanding and route planning tasks, experimental results indicate that there remains a significant gap between the models' visual reasoning capabilities and human cognitive levels. To address this core issue, this study proposes a topological representation method that integrates room features and path information, constructs a navigation-friendly semantic priori map, and enhances VLMs' ability to understand spatial structures.

On the basis of obtaining the semantic priori map, this study adopts a hierarchical chain-of-thought prompting template and performs joint output in combination with the annotated priori maps. By defining the starting segment and target room, accurate global navigation plannings can be generated and the navigation planning performance of the VLMs in complex FBs can be greatly improved.

To enable intelligent agents to generate precise actions in complex FBs, this paper proposes a multi-model collaborative mechanism for outputting action instructions required for localization and navigation planning execution. Specifically, the agent completes self-localization by identifying surrounding landmarks before the planning phase. After the planning phase, the landmarks to be detected and their corresponding first-person action sequences are clarified according to the planning results. Among them, coarse-grained actions are generated by VLMs, while fine-grained actions are generated through the collaboration of the base vision model and the PixelNav neural network\cite{cai2024bridging}. This mechanism forms a closed-loop process of "target determination → search → verification → target update", ensuring that the agent can continuously advance toward the final target. Given the fact that no suitable simulation environment exists for FBs now, we also construct 6 FB simulation environments with distinct structures based on the Gazebo platform. In summary, the main contributions of this paper are as follows:

\begin{itemize}
\item{A priori map guided embodied navigation framework PM-Nav is proposed to mimic human navigation behaviors in FB scenarios.}
\item{In the PM-Nav, environmental maps are parsed into segment-based semantic ones to enhance  the logical reasoning limitation of the VLMs, a H-CoT prompting template is introduced to integrate priori-map cues with symbolic annotations to generate accurate global navigation plans, and a multi-modal collaborative mechanism is designed to produce fine-grained actions. To the best of our knowledge, it is the first work to deeply explore embodied navigation performance in FBs.}
\item{A novel navigation dataset is constructed to facilitate navigation research for FBs.}
\item{Both simulation and real-world experiments show that the PM-Nav outperforms existing zero-shot navigation SOTA methods greatly.}
\end{itemize}

\section{RELATED WORK}
\subsection{Simulators for Embodied Navigation} Current research on embodied navigation mainly focuses on three environments, i.e., indoor household navigation (R2R \cite{anderson2018vision}, R4R \cite{jain2019stay}, RxR \cite{ku2020room}) based on discrete settings like Matterport3D \cite{chang2017matterport3d}, continuous navigation like VLN-CE \cite{krantz2020beyond} using platforms such as Habitat \cite{savva2019habitat}, and outdoor navigation (Touchdown \cite{chen2019touchdown}, Map2Seq \cite{schumann2022analyzing}) based on urban street views. Although these studies cover diverse environments such as homes, offices, and city streets, research on functional buildings (FBs) with complex layouts and densely distributed and highly similar landmarks is not available yet.

\subsection{Path Planning Methodologies} Several path planning methods exist for various navigation tasks. Traditional approaches partially rely on predefined topological structures of the environment, performing graph-based path search to improve planning \cite{patel2024stage}. Alternatively, reinforcement learning and imitation learning methods focus on acquiring navigation policies from experience to realize end-to-end VLN \cite{liu2024vision}. Approaches based on semantic maps attempt to construct visual-semantic representations and leverage graph neural networks for navigation decisions, enhancing the coordination between perception and control \cite{shan2025graph2nav}. Recently, VLM-assisted methods aim to exploit their powerful reasoning capabilities for task planning and semantic annotation for end-to-end VLN in unseen scenes\cite{chen2025affordances}. In this work, we propose a novel path planning approach that integrates structured semantic representations with the reasoning abilities of VLMs, improving VLMs' capability in understanding environment maps and generating effective first-person navigation plannings.

\subsection{VLMs based Embodied Navigation}
VLMs are increasingly applied to embodied navigation tasks due to their reasoning capabilities. MapGPT \cite{chen2024mapgpt} leverages map-guided GPT agents for global planning. Mobility VLA \cite{chiang2024mobility} combines VLMs' contextual reasoning with robust graph-based navigation. PixelNav \cite{cai2024bridging} uses pixels as goal specifications. NavGPT \cite{zhou2024navgpt} handles multimodal input, open-world interaction, and progress tracking. DiscussNav \cite{long2024discuss} enables expert-based collaborative decision-making. Despite their success, they lack the ability to navigate in FBs where objects have highly similar features. In contrast, we design an environment map parsing and multi-model collaborative action output method to help agents understand the spatial relationship between paths and rooms, and locate landmarks within the environment.

\section{METHODOLOGY}
\begin{figure*}[t]
    \centering
    \includegraphics[width=\textwidth]{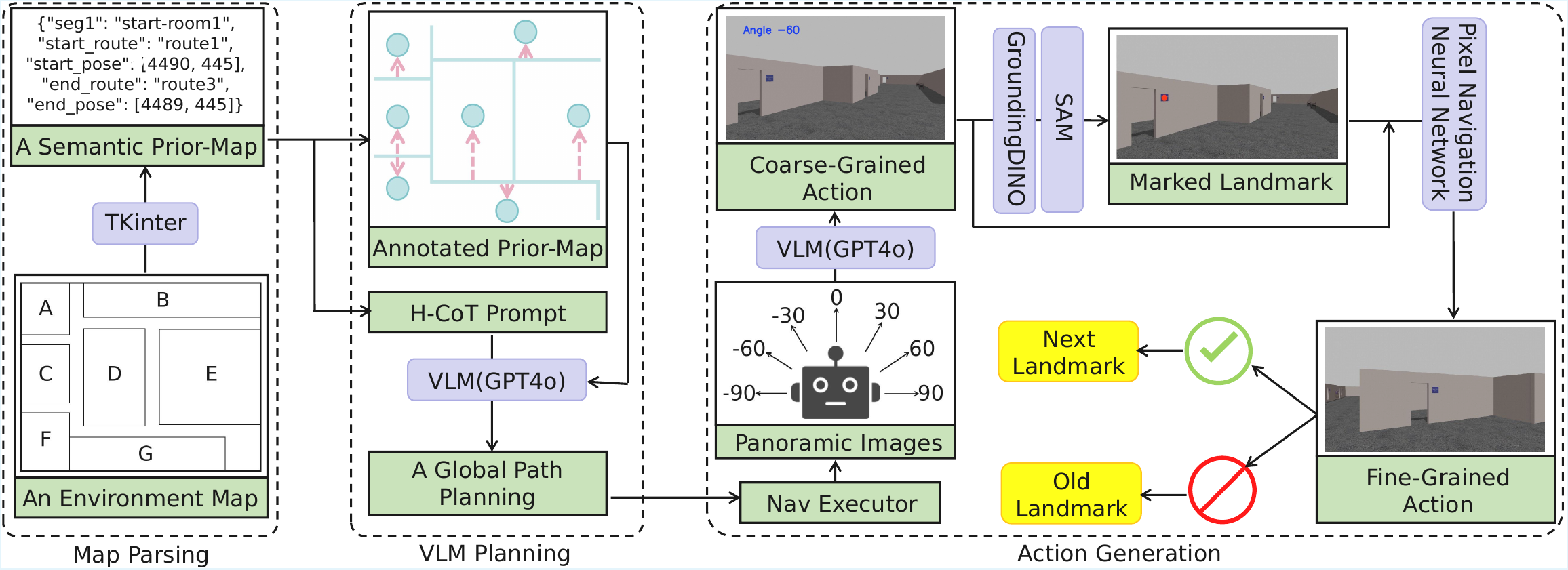}
    \caption{Overview of the Priori-Map Guided Embodied Navigation.}
    \label{fig1}
\end{figure*}
Fig. \ref{fig1} illustrates the proposed framework. Firstly, a priori environment map is parsed into a semantic priori one that encodes associative relationships between paths and rooms. Secondly, a hierarchical chain-of-thought prompting (H-CoT prompt) template is combined with the annotated priori-map and provided to a VLM to generate the path plannings that guide subsequent exploration, and the target is identified based on the Nav Executor. Finally, a panoramic image of the current environment is captured and input to the VLM to generate coarse-grained actions. Subsequently, GroundingDINO and SAM work in collaboration to mark the target landmark. The coarse-grained action information, combined with the marked landmark, is then fed into the Pixel Navigation Neural Network to generate fine-grained actions. If the target is successfully detected, the system proceeds to detect the next landmark. Otherwise, it continues to locate old landmark.

\subsection{Map Parsing}
Since VLMs excel in textual reasoning \cite{mitchell2023comparing} and structured text aids inference \cite{rana2023sayplan}, we parse the environment map to a structured semantic priori-map, using a Tkinter interface\cite{tkinter_python_doc}. Due to the common presence of the high similarity in FB environments, we assign room IDs for efficient internal communication. For paths, we define key waypoints, e.g., start, turn, branch, and end fork, and use these annotations to construct the path structure. To reduce the cognitive load on the VLM, we design a segmented representation combining room and path information, effectively representing the room-path topology with fewer details. Room waypoints are linked to the nearest room based on distance, and the path is divided into segments based on the key waypoints and the room waypoints, such as seg13(room14–room7), which indicates that in the path, segment 13 is located between room14 and room7. This process is executed once per environment. By using this method, we simplify environment map representation into structured semantic priori-map, enabling VLMs to leverage their textual reasoning capabilities for navigation plannings.

\subsection{VLM Planning}
Although the semantic priori-maps can reduce the difficulty of map understanding for VLMs to some extent, they still have limitations in generating accurate path plannings. To further improve VLM navigation plannings, we design a H-CoT prompt template which includes descriptions of these visual tags to strengthen the model's environmental understanding and path-planning accuracy. Meanwhile, since tokenized image information is proven to enhance VLM output quality\cite{yang2023set}, we construct an annotation priori-map based on the semantic priori-map to assist navigation planning.

The navigation task is divided into three steps, i.e., determining the relative spatial relationship between the agent's current position and the target on the priori-map, identifying key waypoints required to reach the destination, and generating corresponding first-person navigation actions based on the waypoint information. We additionally structure the spatial relationships between key path segments, e.g., "from start to turn1, the direction is south", to help the VLM infer the correct actions at decision points, thereby improving navigation accuracy. Fig. \ref{fig2} presents details of the global path planning output by the VLM, where the planning starts at seg13 and targets the room17 area. This path planning takes room nodes and special marker points as the division basis, decomposing the overall path into several consecutive road segments. Within this framework, the navigation task of the agent for each road segment is simplified to locate and reach the target landmark corresponding to that road segment, without the need to process global path information.

\subsection{Action Generation} 
To support general-purpose environment exploration and ensure accurate localization of target landmarks, we propose a collaborative framework that integrates VLM, base vision models, and PixelNav neural network \cite{cai2024bridging}. The agent first uses the VLM to generate coarse-grained action commands. Specifically, the robot rotates to capture images from six directions, which are stitched into a panoramic view with annotated observation angles. During initial exploration, a target is predefined, and the VLM is prompted to locate the landmark within the panorama images and predict its direction. This enhances the model’s spatial reasoning. Since the initial outputs are coarse, e.g., every 30 degrees, we refine them using a combination of GroundingDINO \cite{liu2024grounding}, SAM \cite{kirillov2023segment} and neural networks. The mask center, along with the current first-person observation, is fed into a neural network based on a fine-tuned PixelNav structure, which outputs a more precise direction toward the target.

\begin{figure}[!t]
  \centering
  \includegraphics[width=\columnwidth]{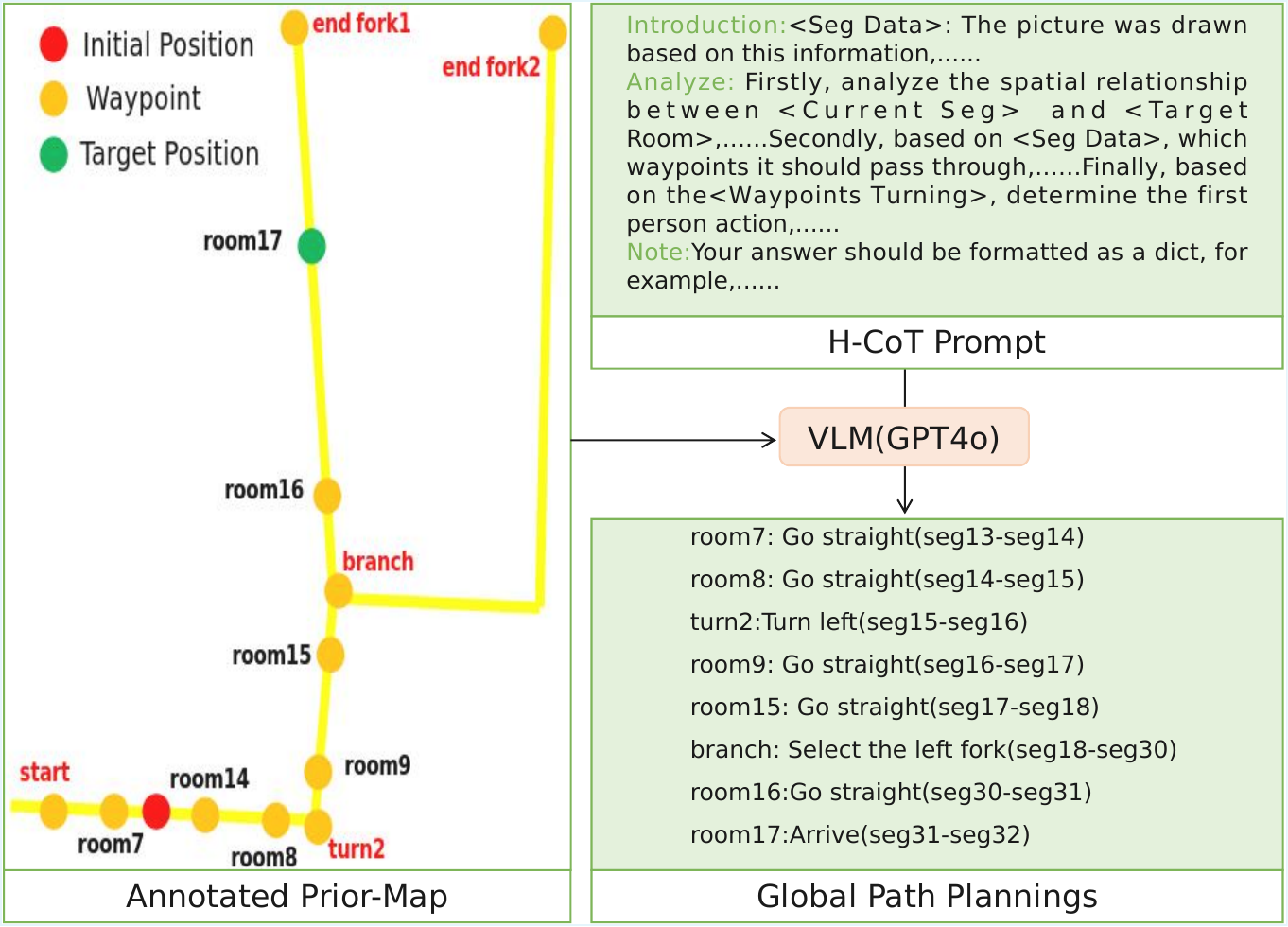}
  \caption{Details of the VLM Planning model.}
  \label{fig2}
\end{figure}
\textbf{Localization} The agent captures images at 30°, 90°, 150°, 210°, 270°, and 330° to explore nearby landmarks. At this stage, no specific target is required and identifying any two landmarks is sufficient to determine the current position. If fewer than two landmarks are visible, the agent continues exploring along the path.
\begin{figure*}[t]
    \centering
    \includegraphics[width=\textwidth]{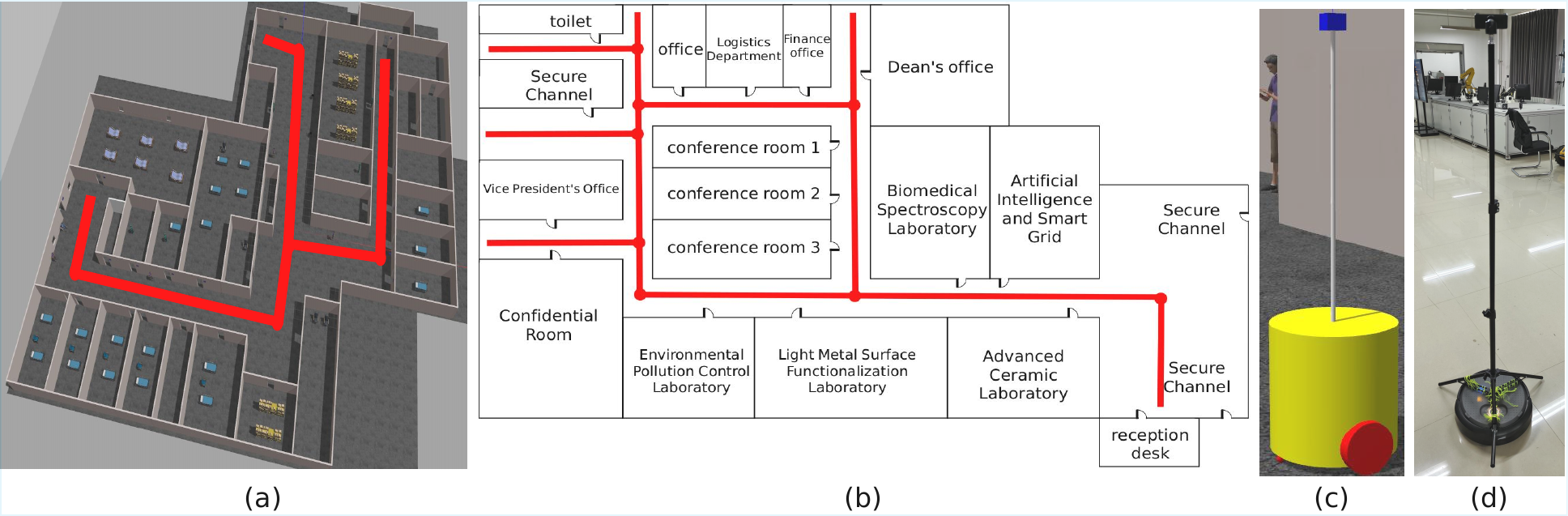}
    \caption{(a)A top-down view of one of the simulation scenarios. (b)The environmental map of a real school. (c)a wheeled robot. (d)Irobot created 3.}
    \label{fig3}
\end{figure*}

\textbf{Navigation.} The agent follows a pre-planned path. As illustrated in Fig. \ref{fig2}, if the next landmark is the room 7 and the required action is "Go straight", the agent captures images at 0°, ±30°, ±60°, and ±90° to search for the target. One view is excluded based on the previous action, e.g., if the last action was a left turn, -90° is excluded. Upon detecting the target, the agent updates the next landmark and corresponding action. For special locations that cannot be directly observed, e.g., “turn2”, we introduce an action prediction mechanism that outputs the current and next action simultaneously. For instance:

\[ \{ \text{room8: Go straight} \} \rightarrow \{ \text{turn2: Turn left} \} \]

After detecting the room 8, the agent must decide whether to turn. This is done by analyzing the mask ratio of walls and floor in front of the robot to distinguish between a corridor and a corner. If a corner is detected, the agent executes the turn. In forked path environments, fine-grained control is required. The agent applies the same landmark-based targeting approach by capturing multi-angle observations and specifying the intended direction. The VLM provides a coarse direction, which is then refined using the neural network to produce a precise action angle.


\section{EXPERIMENTS AND DISCUSSIONS}

To verify the navigation capability of the proposed PM-Nav in FBs, this study designs and implements systematic experiments to explore following research questions (RQs):

\textbf{RQ1:} How does the PM-Nav perform in embodied navigation tasks with various difficulty levels?

\textbf{RQ2:} What are the effects of each proposed module on navigation performance?

\textbf{RQ3:} Can the robot efficiently execute the generated navigation plans?

\textbf{RQ4:} How does the PM-Nav perform in real-world scenarios?

\textbf{Environments.} Due to the lack of adaptive FB simulation environments, this study constructs six FB simulation scenarios(Fig. \ref{fig3}a) based on the environment maps of the VA Design Guide dataset \cite{va2010designguide}. By leveraging Gazebo's high-fidelity physical simulation and sensor emulation capabilities as well as RViz's data visualization functions, a tunable testing platform is established to evaluate the performance of robot navigation algorithms across diverse FB spaces. We also test the embodied navigation capability of the PM-Nav in building 3A, Foshan graduate school, Northeastern university, which contains 18 detectable landmarks(Fig. \ref{fig3}b).

\textbf{Robotic Platform.} A self-designed wheeled robot platform (Fig. \ref{fig3}c), with its camera mounted at a height of 1.6 meters and a horizontal field of view of 60 degrees. In the real-school experiments, we select the iRobot Create3 as the experimental carrier(Fig. \ref{fig3}d), with its camera also mounted at a height of 1.6 meters and a horizontal field of view of 60 degrees.

\subsection{RQ1: Powerful object navigation capability in FBs}
We categorize the tasks into three levels, i.e., simple tasks where the object is within the agent's current visual range and reachable within 3–5 steps, medium tasks where the object is invisible and requires the agent to make at least one turn and 6–10 steps to reach and difficult tasks that need multiple turns and fork selections and demand more than 10 steps to complete. We select the SG-Nav\cite{yin2024sg} and the InstructNav\cite{long2024instructnav}, two zero-shot object navigation methods in continuous environments, for comparison, as their object navigation capabilities have been proved in HM-3D\cite{ramakrishnan2021habitat}. For all navigation tasks in FBs, both the SG-Nav and InstructNav show extremely low SR and SPL as shown in TABLE \ref{table1}. In fact, tasks cannot be completed at all for hard tasks when using these two SOTA navigation methods. This is mainly caused by the fact that they explore in the environment without utilizing priori maps and taking specific objects, e.g., chairs, as targets. However, in FBs, our objects are rooms with identical appearances, and they can only be distinguished  based on specific landmark information. In comparison, the PM-Nav offers 633\% and 1600\%, and 389\% and 750\% SR improvements over the SG-Nav and the InstructNav, for easy and medium tasks, respectively. For hard tasks, where navigation cannot be completed at all, the PM-Nav shows a 46\% SR. All these great improvements elucidate the superior navigation capabilities of the PM-Nav. This is primarily attributed to the fact that PM-Nav derives path planning via priori-maps and explores landmark information through action outputs based on multi-model collaboration for localization and object navigation.
\begin{table}[htbp]
    \centering
    \caption{\centering Comparison in FB simulation environments}
    \label{table1}
    \begin{tabularx}{\columnwidth}{>{\hsize=2.9\hsize\centering\arraybackslash}X *{6}{>{\hsize=0.683\hsize\centering\arraybackslash}X}} 
        \toprule
        \multirow{2}{*}{ } & \multicolumn{2}{c}{Easy} & \multicolumn{2}{c}{Medium} & \multicolumn{2}{c}{Hard}\\
        \cmidrule(lr){2-3} \cmidrule(lr){4-5} \cmidrule(lr){6-7} 
        & SR & SPL & SR & SPL & SR & SPL\\
        \midrule
        SG-Nav\cite{yin2024sg}       & 12   & 4.96  & 4    & 1.92  & 0   & 0.00  \\
        InstructNav\cite{long2024instructnav}  & 18   & 5.81  & 8    & 2.80  & 0   & 0.00  \\
        PM-NaV(Ours)       & 88   & 77.40 & 68   & 58.60 & 46  & 36.40 \\
        \bottomrule
    \end{tabularx}
\end{table}
\subsection{RQ2: All three modules play important roles}
\textbf{Evaluation of Navigation Planning Capability.} We compare three navigation planning scenarios, i.e., inputs using ordinary prompt with environmental maps (O-EM), inputs using H-CoT prompt with environmental maps (H-EM), and inputs using H-CoT prompt with annotated priori-maps (H-PM), and test them using environmental maps from the VA Design Guide dataset \cite{va2010designguide}. TABLE \ref{table2} shows that in simple, medium and hard tasks, O-EM yields extremely low SR, whereas H-EM and H-PM improve SR by over 7 folds. This demonstrates that the H-CoT prompts and annotated priori-maps tremendously promote embodied navigation path planning generation, especially in complex FB navigation tasks.
\begin{figure*}[t]
    \centering
    \includegraphics[width=\textwidth]{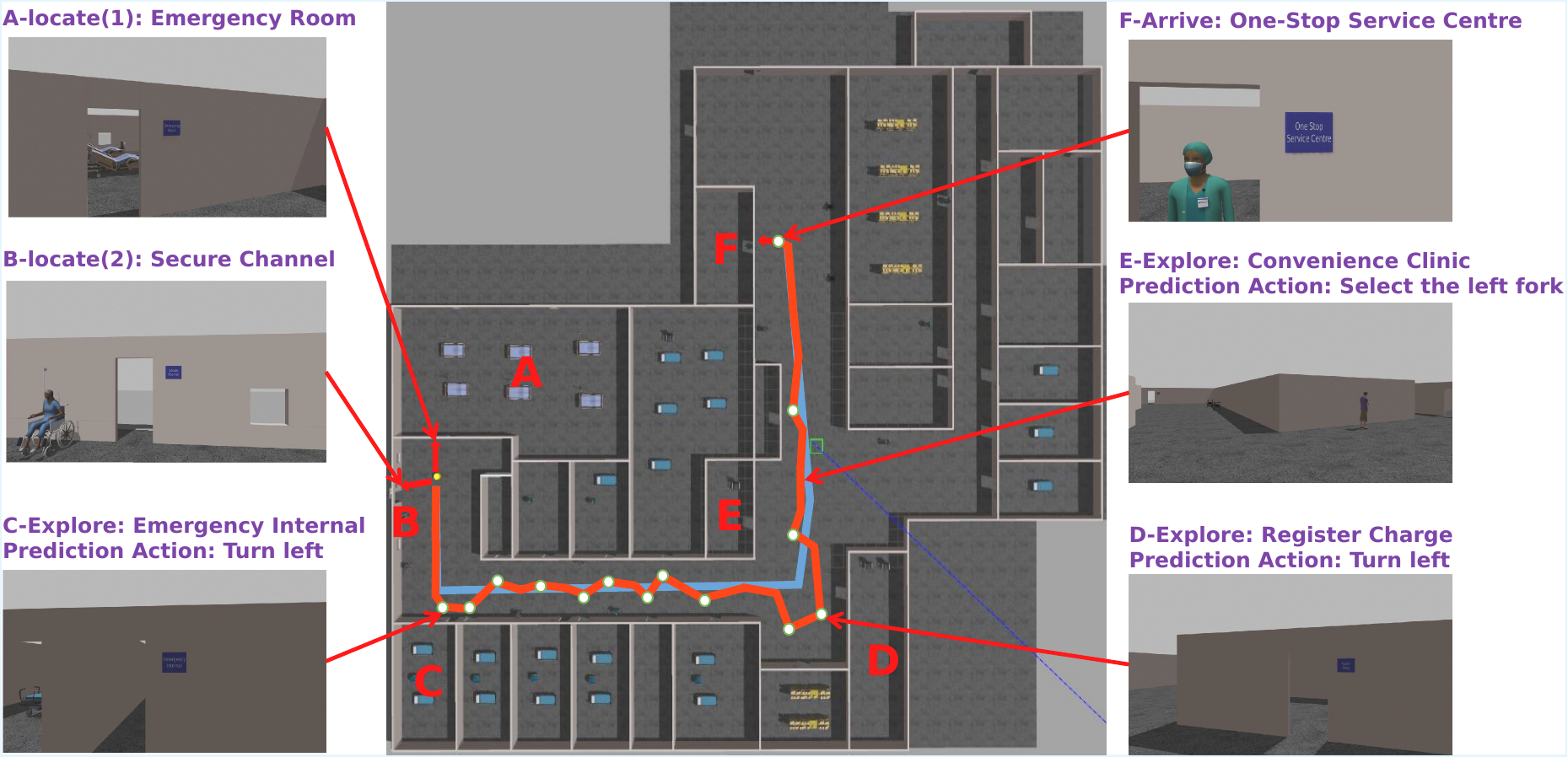}
    \caption{Visualization Comparison of Planned Path and Actual Navigation Trajectory}
    \label{fig4}
\end{figure*}

\begin{table}[htbp]
    \centering
    \caption{\centering Effect of Input Types on VLM Navigation}
    \label{table2}
    \begin{tabularx}{\columnwidth}{>{\centering\arraybackslash}X >{\centering\arraybackslash}X >{\centering\arraybackslash}X >{\centering\arraybackslash}X}
        \toprule
        Input & Easy & Medium & Hard \\
        \midrule
        O-EM & 31.3 & 10.4 & 0.0 \\
        H-EM& 91.6 & 85.4 & 75.0 \\
        H-PM & 95.8 & 89.5 & 83.3 \\
        \bottomrule
    \end{tabularx}
\end{table}

\textbf{Evaluation of Localization Capability.} To evaluate the impact of landmark density on the positioning accuracy of agents, this study classifies landmark density into three levels, as shown in Table \ref{table3}. The redundant status means there are more than two landmarks nearby, allowing the agent to independently select any two for positioning. The minimal status indicates exactly two available landmarks, barely meeting the basic positioning needs. The scarce status refers to only 0-1 landmark present, requiring the agent to actively explore to complete positioning. Given that positioning requires at least two landmarks for detections, this study introduces a new evaluation index, i.e., Success weighted by the Detection Frequency (SDF), which is defined as:
\begin{equation}
\text{SDF} = \frac{1}{N} \sum_{i=1}^{N} S_i \cdot \frac{d_i}{\max(2, d_i)},
\end{equation}
where $S_i = 1$ if localization is successful, and $S_i = 0$ otherwise. $d_i$ represents the number of exploration attempts made by the robot, and $N$ is the total number of trials. The localization SR exceeds 90\% in both the redundant and the minimal status, and 63.3\% in the scarce status. In the redundant status, the agent can choose any two landmarks, allowing the multi-model collaborative action output method to fully leverage its accuracy in identifying surrounding landmarks. In the minimal status, the agent needs to accurately identify exactly two landmarks, due to the uncertainty in the VLM outputs, there is a small chance of repeatedly recognizing the same landmark. This increases the number of recognition attempts. In the scarce state, the agent must explore the environment, significantly increasing the number of exploration steps, leading to a decrease in the SR and a severe reduction in the SDF.

\begin{table}[htbp]
    \centering
    \caption{\centering Effect of Input Types on the PM-Nav}
    \label{table3}
    \begin{tabularx}{\columnwidth}{>{\centering\arraybackslash}X >{\centering\arraybackslash}X >{\centering\arraybackslash}X >{\centering\arraybackslash}X}
        \toprule
        Input & Redundant & Minimal & Scarce \\
        \midrule
        SR & 93.8 & 91.7 & 63.3 \\
        SDF& 90.4 & 82.6 & 25.0 \\
        \bottomrule
    \end{tabularx}
\end{table}

\textbf{Ablation Study on Action Refinement for Robust Navigation.} This study proposes a multi-model action generation strategy, where coarse-grained actions are generated by a VLM, and then refined through a base model-neural network joint optimization module to output fine-grained actions. As shown in TABLE \ref{table4}, the absence of the action optimization module leads to a significant decrease in the agent's navigation success rate for target reaching tasks, verifying the necessity of this fine-grained action process to enhance navigation performance in complex FB scenes.
\begin{table}[htbp]
    \centering
    \caption{\centering Performance on Different Action Generation Levels}
    \label{table4}
    \begin{tabularx}{\columnwidth}{>{\centering\arraybackslash}X *{4}{>{\centering\arraybackslash}X}}
        \toprule
        \multirow{2}{*}{ } & \multicolumn{2}{c}{Coarse-Grained Action} & \multicolumn{2}{c}{Fine-Grained Action} \\
        \cmidrule(lr){1-5}
        & SR & SPL & SR & SPL \\
        \midrule
        Easy & 32 & 22.61 & 88 & 77.4 \\
        Medium & 10 & 7.71 & 68 & 58.6 \\
        Hard & 0 & 0 & 46 & 36.4 \\
        \bottomrule
    \end{tabularx}
\end{table}

\subsection{RQ3: Visualization of Robot Execution Capability}
This section analyzes the robot's execution capability for planning strategies through visual comparison between path planning trajectories and actual navigation trajectories. As shown in Fig. \ref{fig4}, we select a hard object navigation task, and the object is the One-Stop Service Centre. The blue curve represents the global path planning trajectory, and the red curve represents the actual running trajectory of the robot. Each node in the figure marks the position where the robot detects landmarks, and embeds first-person perspective screenshots of key nodes, including Positioning Point A, Positioning Point B, Left Turn Action Prediction C, Left Turn Action Prediction D, Left Fork Selection E, and Target Arrival F.  

Compared with the path planning trajectories and the actual navigation trajectories, the PM-Nav trajectory exhibits significant segmental characteristics, which are strongly correlated with the distribution of landmarks in the path. When observable landmarks exist in the path, the agent demonstrates obvious landmark-approaching behavior, causing the actual trajectory to deviate toward the landmark position. While in areas with no sparse landmarks, the robot travels straight along the planned path without target-oriented path deviations.

\subsection{RQ4: It demonstrates outstanding performance in real-school tests.}
TABLE \ref{table5} reveals that although the SG-Nav and the InstructNav demonstrate favorable navigation performance in private residential scenarios\cite{yin2024sg}, \cite{long2024instructnav}, their performance faces significant challenges in complex real-world FB environments. For all navigation tasks in FBs, both the SG-Nav and InstructNav show extremely low SR and SPL. This is mainly caused by the fact that they struggle to capture landmark information when they need to thoroughly explore the environment due to uncertain object locations, especially when the features of different rooms in FBs are highly similar. In contrast, the PM-Nav offers 650\% and 400\% improvements over the SG-Nav and the InstructNav, for easy tasks, respectively.  For medium and hard tasks, where navigation cannot be completed at all using existing SOTA methods, the PM-Nav shows a 55\% and 15\% SR. This is mainly caused by the fact that the PM-Nav can effectively parse environmental maps and accurately understand the topological relationships between rooms and paths. Meanwhile, it generates fine-grained actions through a multi-model collaboration mechanism, thereby providing stable and reliable decision support for the environment exploration process.
\begin{table}[htbp]
    \centering
    \caption{\centering Comparison on real-world FB environments}
    \label{table5}
    \begin{tabularx}{\columnwidth}{>{\hsize=2.9\hsize\centering\arraybackslash}X *{6}{>{\hsize=0.683\hsize\centering\arraybackslash}X}} 
        \toprule
        \multirow{2}{*}{ } & \multicolumn{2}{c}{Easy} & \multicolumn{2}{c}{Medium} & \multicolumn{2}{c}{Hard}\\
        \cmidrule(lr){2-3} \cmidrule(lr){4-5} \cmidrule(lr){6-7} 
        & SR & SPL & SR & SPL & SR & SPL\\
        \midrule
        SG-Nav       & 10   & 4.57  & 0    & 0  & 0   & 0.00  \\
        InstructNav  & 15   & 4.84  & 0    & 0  & 0   & 0.00  \\
        PM-Nav       & 75   & 50.91 & 55   & 37.16 & 15  & 6.99 \\
        \bottomrule
    \end{tabularx}
\end{table}

\section{CONCLUSIONS}

This paper proposes an embodied navigation framework for functional buildings (FBs), drawing inspiration from humans’ collaborative use of priori maps and landmarks in unfamiliar FBs. Specifically, environmental maps are converted into navigation-friendly semantic priori-maps, a hierarchical thought chain prompt template is designed to support precise path planning, and a multi-agent collaborative action output mechanism is constructed to realize positioning decisions and execution control for navigation planning. Both simulation and real-world tests show tremendous performance improvements over SOTA methods.
Future work will focus on two key directions: (1) improving the accuracy of end-to-end environment map parsing, which remains a challenging task, and (2) enhancing exploration efficiency to minimize unnecessary environmental traversal.

\addtolength{\textheight}{-9cm}   





\bibliographystyle{IEEEtran}

\end{document}